%% file: main.tex
\documentclass[10pt,twocolumn,letterpaper]{article}

\usepackage[pagenumbers]{cvpr} 

\input{preamble}

\definecolor{cvprblue}{rgb}{0.21,0.49,0.74}
\usepackage[pagebackref,breaklinks,colorlinks,citecolor=cvprblue]{hyperref}
\usepackage{graphicx}
\usepackage{amsmath}
\usepackage{amssymb}
\usepackage{booktabs}
\usepackage{float}
\usepackage[accsupp]{axessibility} 
\usepackage[title]{appendix}
\usepackage{lipsum}
\usepackage{multirow}

\usepackage{marvosym}
\usepackage[capitalize]{cleveref}
\crefname{section}{Sec.}{Secs.}
\Crefname{section}{Section}{Sections}
\Crefname{table}{Table}{Tables}
\crefname{table}{Tab.}{Tabs.}

\def\paperID{****} 
\def\confName{****}
\def\confYear{****}

\title{ASH: Animatable Gaussian Splats for Efficient and Photoreal Human Rendering}

\author{Haokai Pang\textsuperscript{1,2 \dag} \quad Heming Zhu\textsuperscript{1 \dag} \quad  Adam Kortylewski\textsuperscript{1,3} \quad Christian Theobalt\textsuperscript{1,4} \quad Marc Habermann\textsuperscript{1,4 \Letter} \\
\textsuperscript{1} Max Planck Institute for Informatics, Saarland Informatics Campus\quad \\
\textsuperscript{2} ETH Zürich \quad
\textsuperscript{3} Universität Freiburg\\
\textsuperscript{4} Saarbrücken Research Center for Visual Computing, Interaction and AI \\
{\tt\small \{hpang, hezhu, akortyle, theobalt, mhaberma\}@mpi-inf.mpg.de}
}

\begin{document}

\input{fig/1_teaser}
\maketitle


\input{sec/0_abstract}
\footnote[0]{\dag\ Joint first authors.}
\footnote[1]{\Letter\ Corresponding author.}
\footnote[2]{Project page: \href{https://vcai.mpi-inf.mpg.de/projects/ash}{\color{magenta}{\url{vcai.mpi-inf.mpg.de/projects/ash}}}}

\input{sec/1_intro}
\input{sec/2_related_work}
\input{sec/3_method}
\input{sec/4_results}
\input{sec/6_conclusion}
\input{sec/7_ack}


{
    \small
    \bibliographystyle{ieeenat_fullname}
    \bibliography{main}
}




\clearpage
\setcounter{figure}{0}
\setcounter{table}{0}
\appendix

\maketitle
\thispagestyle{empty}
\appendix

\section{Overview}
In this appendix, we provide more details regarding the following aspects: 
More implementation details (Sec.~\ref{suppl:ddc}); 
more ablative studies (Sec.~\ref{suppl:ablations});
more results with driven poses from novel datasets (Sec.~\ref{suppl:results});
runtime analysis for the major components (Sec.~\ref{suppl:runtime});
realtime applications built upon \OURS (Sec.~\ref{suppl:application});
more detailed discussion on limitations and future directions (Sec.~\ref{suppl:limitations}).

\input{suppl_sec/1_ddc}
\input{suppl_sec/2_ablations}
\input{suppl_sec/3_results}
\input{suppl_sec/4_runtime}
\input{suppl_sec/6_application}
\input{suppl_sec/5_limitations}

\end{document}

%% file: preamble.tex
\usepackage[dvipsnames]{xcolor}
\usepackage{colortbl}

\newcommand{\OURS}{ASH\xspace}
\usepackage{soul}

\usepackage{pifont}
\newcommand{\good}{{\color{black}\ding{51}}}
\newcommand{\bad}{{\color{black}\ding{55}}}

\definecolor{antiquebrass}{rgb}{0.8, 0.58, 0.46}
\definecolor{amber}{rgb}{1.0, 0.75, 0.0}

%% file: fig/1_teaser.tex
\twocolumn[{%
\renewcommand\twocolumn[1][]{#1}%
\maketitle
\begin{center}
    \centering
    \vspace{-24pt}
    \includegraphics[width=0.95\linewidth]{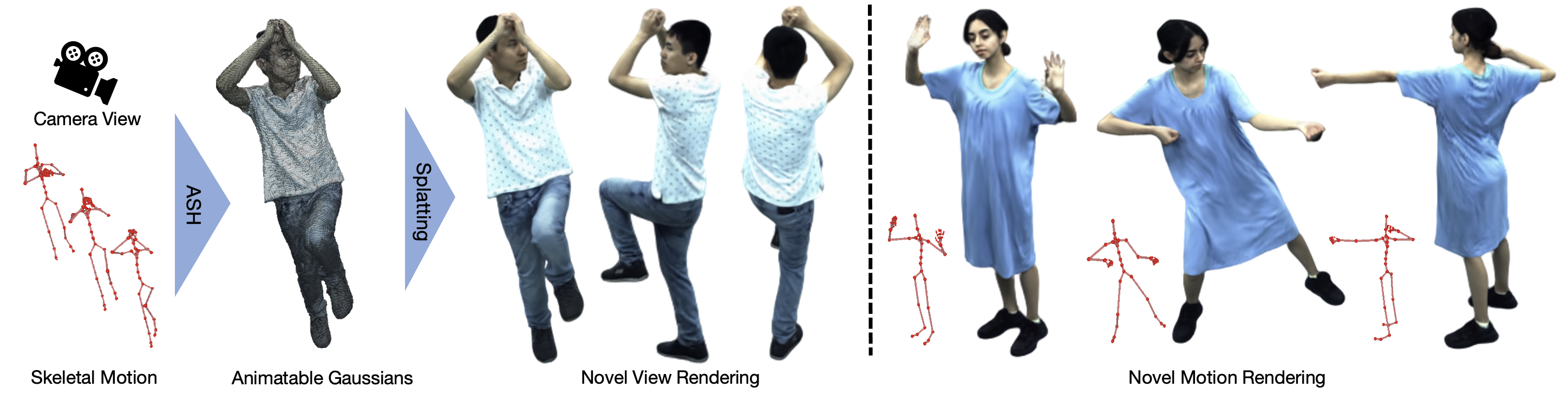}
    \captionof{figure}{
        ASH takes an arbitrary 3D skeletal pose and virtual camera view, which can be controlled by the user, as input, and generates a photorealistic rendering of the human in real time. 
        To achieve this, we propose an efficient and animatable Gaussian representation, which is parameterized on the surface of a deformable template mesh.
    }
    \label{fig:teaser}
    \vspace{-5pt}
\end{center}
}]

%% file: sec/0_abstract.tex
\begin{abstract}
Real-time rendering of photorealistic and controllable human avatars stands as a cornerstone in Computer Vision and Graphics.
While recent advances in neural implicit rendering have unlocked unprecedented photorealism for digital avatars, real-time performance has mostly been demonstrated for static scenes only.
To address this, we propose \OURS, an \textbf{a}nimatable Gaussian \textbf{s}platting approach for photorealistic rendering of dynamic \textbf{h}umans in real time. 
We parameterize the clothed human as animatable 3D Gaussians, which can be efficiently splatted into image space to generate the final rendering.
However, naively learning the Gaussian parameters in 3D space poses a severe challenge in terms of compute.
Instead, we attach the Gaussians onto a deformable character model, and learn their parameters in 2D texture space, which allows leveraging efficient 2D convolutional architectures that easily scale with the required number of Gaussians.
We benchmark \OURS with competing methods on pose-controllable avatars, demonstrating that our method outperforms existing real-time methods by a large margin and shows comparable or even better results than offline methods.
\end{abstract}

%% file: sec/1_intro.tex
%
%
\section{Introduction} \label{sec:intro}
Generating high-fidelity human renderings is a long-standing problem in the field of Computer Graphics and Vision, with a multitude of real-world applications, such as gaming, film production, and AR/VR. 
Typically, this process is a laborious task, requiring complicated hardware setups and tremendous efforts from skilled artists. 
To ease the extensive manual efforts, recent advances, including this work, focus on generating photorealistic and controllable human avatars solely from multi-view videos. 
\par 
Recent works on photorealistic human rendering can be categorized into explicit-based and hybrid methods.
Explicit methods represent the human avatar as a deformable template mesh with learned dynamic textures~\cite{shysheya2019textured,habermann2021}.
Although these methods are runtime-efficient and can be seamlessly integrated with the well-established rasterization-based rendering pipeline, the generated rendering often falls short in terms of photorealism and level of detail.
Hybrid approaches usually attach a neural radiance field (NeRF)~\cite{mildenhall2020nerf} onto a (deformable) human model~\cite{peng2021animatable,liu2020neural,habermann2023hdhumans}.
Typically, they evaluate the NeRF in an unposed space to model the detailed appearance of clothed humans, and generate color and density values by querying a coordinate-based MLP per ray sample.
Although hybrid methods can deliver superior rendering quality through NeRF's capability to capture delicate appearance details, they are unsuitable for real-time applications due to the intensive sampling and MLP evaluations required for volume rendering.
\par 
Recently, 3D Gaussian splatting~\cite{kerbl20233d} with its impressive rendering quality and real-time capability, has become a promising alternative to NeRFs, which are parameterized with a coordinate-based MLP.
However, it originally is only designed for modeling static scenes, which is in stark contrast to our problem setting, i.e., modeling dynamic and animatable human avatars.
Thus, one may ask: Can the rendering quality and speed of Gaussian splatting be leveraged to model the skeletal motion-dependent characteristics of clothed humans, and how can pose control be achieved?
\par 
To answer this, we propose \OURS, a real-time approach for generating photorealistic renderings of animatable human avatars.
Given a skeletal motion and a virtual camera view, \OURS produces photorealistic renderings of clothed humans with motion-dependent details in real time (see Fig.~\ref{fig:teaser}). 
Importantly, during training, \OURS only requires multi-view videos for supervision.
\par 
In more detail, our animatable human avatar is parameterized using Gaussian splats.
However, naively learning a mapping from skeletal pose to Gaussian parameters in 3D leads to inferior quality when constraining ourselves to real-time performance.
Thus, we propose to attach the Gaussians onto a deformable mesh template of the human.
Given the mesh's uv parameterization, it allows learning the Gaussian parameters efficiently in 2D texture space.
Here, each texel covered by a triangle represents a Gaussian.
Thus, the number of Gaussians remains \textit{constant}, which is in stark contrast to the original formulation.
Similarly, we encode the skeletal motion as pose-dependent normal maps.
As a result, learning the mapping from skeletal motion to dynamic and controllable Gaussian parameters simplifies to a 2D-to-2D image translation task, which can be efficiently implemented using 2D convolutional architectures.
For supervision, we transform the Gaussians into global 3D space using the deformable template and learned Gaussian displacements, splat the Gaussians following the original formulation, and supervise solely on multi-view videos.
Our contributions are:
\begin{itemize}
    \item We propose a novel method, \OURS, that enables real-time and high-quality rendering of animatable clothed human avatars solely learned from multi-view video.
    \item To this end, we represent the human avatar as animatable and dynamic Gaussian splats, which we attach to a deformable template.
    \item To efficiently learn such a representation, we phrase the problem as a 2D-to-2D texture translation task effectively circumventing 3D architectures, which do not easily scale to the typically required large number of Gaussians. 
\end{itemize}
Our evaluations and comparisons against state-of-the-art methods on animatable human rendering demonstrate that \OURS is a significant step towards real-time, high-fidelity, and controllable human avatars.

%% file: sec/2_related_work.tex
%
%
\section{Related Work} \label{sec:rw}
\paragraph{Neural Rendering and Scene Representation.} 
%
\par 
In the last few years, volumetric representations~\cite{henzler2019escaping,sitzmann2019deepvoxels,sitzmann2019srns} and neural radiance fields (NeRF)~\cite{mildenhall2020nerf,zhang2020nerf++} have received significant attention due to their ability to generate high-quality geometry and appearance~\cite{tewari2022advances,xie2022neural}.
However, rendering a NeRF is typically slow as it requires querying an MLP for each ray sample during volume rendering. 
To address this, subsequent research focused on accelerating the inference process of NeRF:
Neural Sparse Voxel Fields~\cite{liu2020neural} adopts an octree to prune the ray samples. 
DVGO~\cite{sundvgo} models the scene with an explicit density and feature grid.
Plenoxels~\cite{fridovich2022plenoxels} and PlenOctree~\cite{yu2021plenoctrees} replace the MLP with a hierarchical 3D grid storing spherical harmonics, achieving an interactive test-time framerate. 
TensoRF~\cite{chen2022tensorf} and Instant-NGP~\cite{muller2022instant} achieved faster inference with compact scene representations, i.e., decomposed tensors and neural hash grids.
3D Gaussian Splatting~\cite{kerbl20233d} encodes the scene with Gaussian splats storing the density and spherical harmonics, which achieves state-of-the-art rendering quality and shows real-time capability.
However, all the above methods are tailored for static scenes, and it is non-trivial to extend them for modeling the dynamic appearances of clothed humans.
%
%
\par
There are also notable advancements for extending the concept of NeRFs~\cite{tretschk2020nonrigid,park2020nerfies, pumarola2020dnerf,park2021hypernerf,Li2020NeuralSF,xian2020spacetime} to dynamic scenes. 
However, most of these works only support playback of the same dynamic sequence under novel views and, therefore, cannot be adopted for user-controlled pose-dependent dynamic appearance of clothed humans.
\paragraph{Animatable Neural Human Rendering.}
Since this work focuses on animatable human rendering, i.e., at test time, the approach solely takes the skeletal motion as input, we do not discuss works on replay~\cite{peng2021neuralbody,Lombardi2021MVP,wang2020learning,kwon2021neural,weng_humannerf_2022_cvpr,isik2023humanrf}, reconstruction~\cite{xiang2021modeling,alldieck18b,alldieck19,POP:ICCV:2021,Saito:SCANimate:2021,tiwari21neuralgif,LEAP:CVPR:21,lin2022fite,li2020monocular,jiang2022hifecap,habermann2021deeper,habermann2020deepcap,habermann2019livecap}, and image-based free-viewpoint rendering~\cite{wang2021ibrnet,Remelli2022TexelAligned,shetty2023holoported}. 
Here, according to the underlying shape representation and rendering scheme, we can categorize the literature into two streams, i.e., mesh-based methods and hybrid methods.
%
%
\par 
Mesh-based methods~\cite{xu2011video, casas14, Volino2014, shysheya2019textured, habermann2021, xiang2022dressing, bagautdinov2021driving} adopt an explicit, motion-controllable template mesh to model the geometry of clothed humans, with texture space for encoding appearance features.
Xu et al.~\cite{xu2011video} first achieved novel motion and pose synthesis by querying and wrapping texture patches from the captured dataset. 
Casas et al.~\cite{casas14} and Volino et al.~\cite{Volino2014} proposed an interactive system that models the appearance as a temporally consistent layered representation in textures space. 
However, the rendering quality is limited due to the coarse geometry proxy. 
TNA~\cite{shysheya2019textured} adopts a texture stack for modeling the dynamic humans' appearances, though it cannot generate motion-dependent appearance. 
To address this issue, DDC~\cite{habermann2021} employs differentiable rendering to learn the non-rigid deformations and dynamic texture maps of clothed humans. 
At test time, DDC can generalize to novel poses and views and produce real-time photorealistic renderings. 
Our method outperforms DDC in terms of rendering quality by a large margin while maintaining real-time capability.
%
%
\par 
Although mesh-based methods provide intuitive control through skeletal poses and integrate seamlessly with established rasterization pipelines, their rendering quality is restrained by the resolution of the template mesh.
To this end, hybrid methods are introduced, which articulate the implicit fields with the explicit shape proxy, i.e., parametric human body models~\cite{loper15,STAR:2020,SMPL-X:2019,TotalCapture2018}, or person-specific template meshes. 
A popular line of research~\cite{su2021nerf,weng2020vid2actor,chen2021animatable, 2021narf,ARAH,bergman2022generative,jiang2022instantavatar,li2022tava,su2022danbo,Feng2022scarf,SHERF} introduced deformable human NeRFs that unwrap the posed space to a shared canonicalized space with inverse kinematics. 
To better model the pose-dependent appearance of humans, recent studies ~\cite{liu2021neural, peng2021animatable, xu2021hnerf, NNA, zheng2023avatarrex, habermann2023hdhumans, kwon2023deliffas,zhu2023trihuman} further introduce motion-aware residual deformations in the canonicalized space.
Neural Actor~\cite{liu2021neural} and HDHumans~\cite{habermann2023hdhumans} are most closely related to our work within this category.
Neural Actor utilizes the texture map of the parametric human body mesh as local pose features to infer dynamic appearances.
However, it fails to generalize to characters with loose outfits. 
HDHumans jointly optimizes the neural implicit fields and the explicit template mesh and, thus, is able to handle loose clothing.
However, both methods are slow and take roughly 5 seconds to render a single frame.
In stark contrast, our proposed method is capable of real-time rendering with a quality on par with or even superior to HDHumans.
%
%

%% file: sec/3_method.tex
%
%
\section{Method} \label{sec:method}
Our goal is to generate motion-controllable, photorealistic renderings of humans learned solely from multi-view RGB videos (Fig.~\ref{fig:pipline}).
Specifically, \OURS takes the skeletal motions and a virtual camera view as input at inference and produces high-fidelity renderings in real-time ($\sim$$\mathbf{30fps}$).
To this end, we propose to model the dynamic character with 3D Gaussian splats, parametrized as texels in the texture space of a deformable template mesh.
This texel-based parameterization of 3D Gaussian splats enables us to model the mapping from skeletal motions to the Gaussian splat parameters as a 2D image-2-image translation task.
Next, we will explain \OURS from the following aspects: 
The background and problem setting (Sec.~\ref{sec:settings}), modeling animatable Gaussian splats (Sec.~\ref{sec:anigaus}), and our dedicated training strategy tailored towards our animatable Gaussians (Sec.~\ref{sec:supervision}).
\input{fig/3_pipeline}
%
%
\subsection{Problem Setting and Background} \label{sec:settings}
We assume a segmented multi-view video $\mathbf{I}_{f,c}$ of an actor, recorded in a studio with $C$ synchronized and calibrated cameras, where $f$ and $c$ denote the frame and camera IDs, respectively.
$\mathbf{C}_c$ denotes the camera projection matrix.
Additionally, each frame $\mathbf{I}_{f,c}$ is annotated with the 3D skeletal pose $\boldsymbol{\theta}_f \in \mathbb{R}^D$ using a markerless motion capture system~\cite{captury}. 
Here, $D$ indicates the number of degrees of freedom (DoFs) for the character's skeleton.
The skeletal motion of the subject $\boldsymbol{\bar{\theta}}_{f} \in \mathbb{R}^{k \times D}$ 
is depicted by a sliding window of skeletal poses from frame $f - k + 1$ to frame $f$ where the root translation is normalized w.r.t. the $f$th frame.
%
%
\par 
For training, our model takes the skeletal poses $\boldsymbol{\theta}_f$ and camera parameters $\mathbf{C}_c$ as input, renders the animatible Gaussians into image space, and is supervised solely on the multi-view video $\mathbf{I}_{f,c}$.
During inference, \OURS takes arbitrary skeletal poses $\boldsymbol{\theta}'$ and virtual cameras $\mathbf{C}'$ as input and generates photorealistic rendering of the subjects at real-time frame rates ($\mathbf{29.64fps}$). The detailed runtime breakdown is reported in the appendix.
%
%
\par 
\noindent\textbf{Gaussian Splatting.}
We paramterize the actor representation as 3D Gaussians, which has been proven to be an efficient representation for modeling and rendering static 3D scenes~\cite{kerbl20233d}.
Here, the static scene is depicted as a collection of 3D Gaussians
%
\begin{equation}
    G(\mathbf{x}) = e^{-\frac{1}{2} (\mathbf{x})^T \Sigma^{-1} (\mathbf{x})}
\end{equation}
%
where $\Sigma$ denotes the covariance matrix and the Gaussian is centered at $\boldsymbol{\mu}$.
In Kerbl et al.~\cite{kerbl20233d}, the Gaussians are parameterized with the set $\mathcal{G}_i = (\boldsymbol{\mu}_i, \mathbf{q}_i, \mathbf{s}_i, \mathbf{\alpha}_i, \boldsymbol{\eta}_i)$, each defined by its position $\boldsymbol{\mu}_i \in \mathbb{R}^{3}$, rotation quaternion $\mathbf{q}_i \in \mathbb{R}^{4}$, scaling $\mathbf{s}_i \in \mathbb{R}^{3}$, opacity $\mathbf{\alpha}_i \in \mathbb{R}$, and spherical harmonics coefficients $\boldsymbol{\eta}_i  \in \mathbb{R}^{48}$.
To render the Gaussians into a particular camera view $c$, the Gaussian $i$ has to be projected into image-space by updating the covariance as
\begin{equation}
\Sigma_{i,c} = \mathbf{J}_c  \mathbf{C}_c \mathbf{R}_i\mathbf{S}_i\mathbf{S}_i^T\mathbf{R}_i^T  \mathbf{C}^T_c \mathbf{J}_c^T 
\end{equation}
where $\mathbf{R}_i$ and $\mathbf{S}_i$ are rotation and scaling matrices obtained from the quaternion convertion $\mathbf{q}_i$ and the scaling coefficients $\mathbf{s}_i$. $\mathbf{J}_c$ is the Jacobian of the affine approximation of the projective
transformation $\mathbf{C}_c$.
%
%
\par
To render the color $\mathbf{c}_\mathbf{p}$ of a pixel $\mathbf{p}$ in camera $c$, 3D Gaussian splatting~\cite{kerbl20233d} adopts a point-based splatting formulation, which blends the spherical harmonics $\boldsymbol{\eta}_i$ of the depth-ordered Gaussian splats overlapping with the pixel as
%
\begin{equation}
	\mathbf{c}_\mathbf{p} = \sum_{j \in \mathcal{N}_p}
	H(\boldsymbol{\eta}_i, \mathbf{d}_\mathbf{p})\alpha^{\prime}_{j}
	\prod_{k=1}^{j-1}(1-\alpha^{\prime}_{k}),
\end{equation}
%
where $\mathcal{N}_p$ denotes the set of Gaussian splats covering pixel $\mathbf{p}$.
$\alpha^{\prime}_{j}$ refers to the opacity for the $j$th ordered Gaussian splat with respect to the current pixel, i.e. $\alpha^{\prime}_{j} = \alpha_jG_j(\mathbf{p})$.
$H(\cdot)$ indicates the function that converts the spherical harmonics coefficients $\boldsymbol{\eta}_i$ and the view direction $\mathbf{d}_{\mathbf{p}}$ to an RGB color.
%
%
\par
While 3D Gaussian splatting can produce high-quality renderings at very high frame rates (more than $100$ fps), its usage is primarily demonstrated for static scenes, and it is non-trivial to adopt this concept for controllable, detailed, and dynamic 3D human avatars. 
What is required here is \textit{animatible} 3D Gaussians, i.e. we want to model the set of Gaussian parameters $\{\mathcal{G}_i\}_{N_g}$ as a function of skeletal motion $\boldsymbol{\bar{\theta}}_{f}$ where $N_g$ denotes the total number of Gaussians.
Note that we consider motion rather than pose to account for potential surface dynamics.
%
%
\subsection{Animatable Gaussian Splats} \label{sec:anigaus}
Intuitively, we want to learn a function $\mathcal{F}(\boldsymbol{\bar{\theta}}_{f}) = \{\mathcal{G}_i\}_{N_g}$ that maps the skeletal motion to animatable 3D Gaussian parameters.
However, more than $20{,}000$ Gaussian splats are typically required to achieve high-fidelity renderings of clothed humans.
Thus, modeling and learning such a function can be challenging, especially when modeling it in 3D.
Instead, our idea is to attach the Gaussian splats onto an animatable template mesh of the human, and parameterize the Gaussian splats in 2D texture space, i.e., each texel of the template mesh (covered by a face) stores the parameters of a 3D Gaussian.
This enables \OURS to efficiently learn the Gaussian parameters in 2D texture space, which we will now describe in more detail.
%
%
\par
\noindent\textbf{Animatable Template.}
To achieve this, we require an animatable human template denoted as $M(\boldsymbol{\theta}_{f})=\mathbf{V}_{f}$, which takes the skeletal motion and computes posed and deformed 3D vertices $\mathbf{V}_{f}$ of a person-specific template mesh $\mathbf{V}_{\mathrm{m}}$.
In practice, we leverage the character model of Habermann et al.~\cite{habermann2021} and refer to the appendix for further details. 
To generate the animatable template mesh $\mathbf{V}_{f}$, we first 
non-rigidly deform the original template mesh vertices $\mathbf{V}_{\mathrm{m}}$ in the unposed-canonical space, denoted as $\bar{\mathbf{V}}_{f}$, with skeletal motion-dependent, i.e. $\bar{\boldsymbol{\theta}}_{f}$, and learned embedded derformations~\cite{embedded} and per-vertex displacements.
Given the skeletal pose ${\boldsymbol{\theta}}_{f}$, the canonically deformed template mesh vertices $\bar{\mathbf{V}}_{f}$ can then be posed using Dual Quaternion skinning~\cite{kavan2007skinning}, denoted as $\mathbf{V}_{f}$.

\noindent\textbf{Animatable Gaussian Textures.} \OURS depicts the character's appearance with a \textit{fixed} number of animatable Gaussian splats $\{\mathcal{G}_{i}\}_N = (\boldsymbol{\bar{\mu}}_{\mathrm{uv},i}, \mathbf{\bar{d}}_{\mathrm{uv},i}, \mathbf{q}_{\mathrm{uv},i}, \mathbf{s}_{\mathrm{uv},i}, \mathbf{\alpha}_{\mathrm{uv},i}, \boldsymbol{\eta}_{\mathrm{uv},i}) \in \mathbb{R}^{N\times62}$ as the texels on the texture space of the animatable template mesh $M(\boldsymbol{\theta}_{f})$.
Here, $N$ denotes the number of texels that are covered by triangles in the UV map.
Specifically, $\boldsymbol{\bar{\mu}}_{\mathrm{uv},i}$ denotes the base position for Gaussian splats in the canonical space, which can be derived from the canonical animatable template mesh vertices $\bar{\mathbf{V}}_{f}$ through texture mapping:
\begin{equation}
    \boldsymbol{\bar{\mu}}_{\mathrm{uv},i} = w_{\mathrm{a},i}\mathbf{\bar{V}}_{f,j} + w_{\mathrm{b},i}\mathbf{\bar{V}}_{f,k} + w_{\mathrm{c},i}\mathbf{\bar{V}}_{f,l},
\end{equation}
where $w_{(\cdot),i}$ denotes the barycentric weights for the texels and $\bar{\mathbf{V}}_{f,(\cdot)}$ stands for the canonical vertex position for the triangle that covers the texel. 
Similar to the animatable template, we can pose the Gaussian splats $\{\mathcal{G}_{i}\}$ stored in texels, from the canonical position $\boldsymbol{\bar{\mu}}_i$ to the posed space, through Dual Quaternion skinning ~\cite{kavan2007skinning}:
\begin{equation}
	\boldsymbol{\mu}_{\mathrm{uv}, i} = \mathbf{T}_{\mathrm{uv}, i}(\boldsymbol{\bar{\mu}}_{\mathrm{uv}, i} + \mathbf{\bar{d}}_{\mathrm{uv}, i}),
\end{equation}
%
where $\mathbf{T}_{\mathrm{uv}, i}$ denotes the Dual Quaternion skinning transformation matrix for the $i$th texel. $\mathbf{\bar{d}}_{\mathrm{uv}, i}$ refers to a learned per-texel offset in the canonical space, which captures fine motion-dependent deformations of the Gaussian splats. 
\par 
Parameterizing Gaussian splats as 2D texels enables us to predict them using efficient 2D convolutional architectures.
Moreover, the shared canonical 2D space facilitates the learning of the motion-dependent Gaussian parameters.
\par
\noindent\textbf{Gaussian Texture Decoder.} 
Due to the texel-based 2D parameterization of the 3D Gaussian splats, we can leverage the well-established, efficient 2D convolutional architectures.
To formulate the mapping between the 3D skeletal motion $\boldsymbol{\bar{\theta}}_{f}$ and the dynamic Gaussian splats $\{\mathcal{G}_i\}_N$ on 2D texture space as a image-2-image translation problem~\cite{pix2pix2017}, 
we adopt the motion-aware textures $(\mathbf{T}_{\mathrm{n},f}, \mathbf{T}_{\mathrm{p},f})$ to depict the 3D skeletal motions $\boldsymbol{\bar{\theta}}_{f}$ in the 2D texture space. 
The normal textures $\mathbf{T}_{\mathrm{n},f}$ and position textures $\mathbf{T}_{\mathrm{p},f}$ can be computed from the posed and deformed template mesh $\mathbf{V}_{f}$ vertices through inverse texture mapping.
Consequently, we propose motion-aware 2D convolutional neural networks, i.e., the geometry network $\mathcal{E}_{\mathrm{geo}}$, and the appearance network $\mathcal{E}_{\mathrm{app}}$, predicting the geometry and appearance parameters of the Gaussian splats from the motion-aware textures $(\mathbf{T}_{\mathrm{n},f}, \mathbf{T}_{\mathrm{p},f})$.
The geometry network $\mathcal{E}_{\mathrm{geo}}$ predicts the shape-related parameters, namely, the canonical offset $\mathbf{\bar{d}}_{\mathrm{uv}, i}$, scale $\mathbf{s}_{\mathrm{uv}, i}$, rotation quaternions $\mathbf{q}_{\mathrm{uv}, i}$, and opacity $\mathbf{\alpha}_{\mathrm{uv}, i}$:
%
\begin{equation} 
\begin{split}
  \mathcal{E}_{\mathrm{geo}}(\mathbf{T}_{\mathrm{n},f}, \mathbf{T}_{\mathrm{p},f}) = (\mathbf{\bar{d}}_{\mathrm{uv}, i}, \mathbf{s}_{\mathrm{uv}, i}, \mathbf{q}_{\mathrm{uv}, i}, \mathbf{\alpha}_{\mathrm{uv}, i}).
\end{split}
\end{equation}
%
\par
A separated motion-aware convolution decoder $\mathcal{E}_{\mathrm{app}}$ is adopted for learning the appearances characterized by the Spherical Harmonics $\boldsymbol{\eta}_{\mathrm{uv}, i}$:
%
\begin{equation} 
\begin{split}
  \mathcal{E}_{\mathrm{app}}(\mathbf{T}_{\mathrm{n},f,}, \mathbf{T}_{\mathrm{p},f},\Phi_f) = \boldsymbol{\eta}_{\mathrm{uv}, i} ,
\end{split}
\end{equation}
%
where $\Phi_f$ indicates the global appearance features, which encodes the global root transition of the character with a shallow MLP, to account for the spatially varying lighting conditions within the capture space.
%
%
\subsection{Training Strategy} \label{sec:supervision}
Unlike static scenes, dynamic clothed humans exhibit motion-dependent appearances and varying geometry throughout the frames, posing a significant challenge in training.
To make it tractable, we propose a carefully designed training paradigm, which decomposes the learning of the motion-aware convolutions into two stages, namely, the warmup stage, and the final training. 
%
%
\par
\noindent\textbf{Warmup Stage.} 
As mentioned in Sec.~\ref{subsec:datset}, the DynaCap dataset~\cite{habermann2021} and our proposed dataset feature long training sequences with various motion-dependent detailed appearances. 
Therefore, naively training the proposed motion-aware decoders $\mathcal{E}_{\mathrm{geo}}$ and $\mathcal{E}_{\mathrm{app}}$, from scratch without proper initialization will not converge during training. 
To tackle this problem, we propose a warmup stage, providing a better weight initialization for the motion-aware decoders.
%
%
\par
We first sample $t$ frames evenly across the training sequence and learn 3D Gaussian splat parameters $\{\mathcal{G}^{\prime\prime}_i\}_{N_\mathrm{g}}$ separately, which serves as a pseudo ground truth for the Gaussian splat parameters.
In contrast to the original implementation for static 3D Gaussian splatting~\cite{kerbl20233d}, we fix the position of the Gaussian splats $\boldsymbol{\mu}^{\prime\prime}_{\mathrm{uv},i}$ 
throughout the training while only optimizing the remaining parameters. 
Specifically, the initial value for the Gaussian splat positions $\boldsymbol{\mu}^{\prime\prime}_{\mathrm{uv},i}$ can be read out from the texture texels of the pose-deformed template mesh $\boldsymbol{\mu}_{\mathrm{uv},i}$, 
Additionally, to preserve the correspondences across pseudo ground truth frames, we remove the splitting/merging of the Gaussian splats and keep the number of Gaussian splats fixed.
The pretraining optimizes the L2 loss between the pseudo ground truth $\{\mathcal{G}^{\prime\prime}_{i}\}$ and the Gaussian splat parameters produced by the motion-aware decoders $\{\mathcal{G}^{\prime}_{i}\}$:
%
\begin{equation} 
\begin{split}
  \mathcal{L}_{\mathrm{pre}} = \mathcal{L}_{2}(\{\mathcal{G}^{\prime}_{i}\}, \{\mathcal{G}^{\prime\prime}_{i}\}).
\end{split}
\end{equation}
%
\par
\noindent\textbf{Final Training.}
After the warmup stage, we can further train the motion-aware decoder on the whole training sequence by minimizing the pixel-wise L1 and structural-similarity-index loss between the generated images $\mathbf{I}^{\prime}_{f,c}$ and the multi-view ground truth images $\mathbf{I}_{f,c}$:
%
\begin{equation} 
\begin{split}
  \mathcal{L}_{\mathrm{main}} = \lambda_{\mathrm{pix}}\mathcal{L}_{1}(\mathbf{I}_{f,c}, \mathbf{I}^{\prime}_{f,c}) + \lambda_{\mathrm{str}}\mathcal{L}_\mathrm{ssim}(\mathbf{I}_{f,c}, \mathbf{I}^{\prime}_{f,c}),
\end{split}
\end{equation}
%
where $\mathcal{L}_\mathrm{ssim}$ denotes the structural similarity index loss~\cite{wang2004image} measuring the structural difference between two images. 
$\lambda_{\mathrm{pix}}$ and $\lambda_{\mathrm{ssim}}$ are set to $0.1$ and $0.9$, respectively.
%
%

%% file: fig/3_pipeline.tex
\begin{figure*}[!t]
\centering
\includegraphics[width=0.98\linewidth]{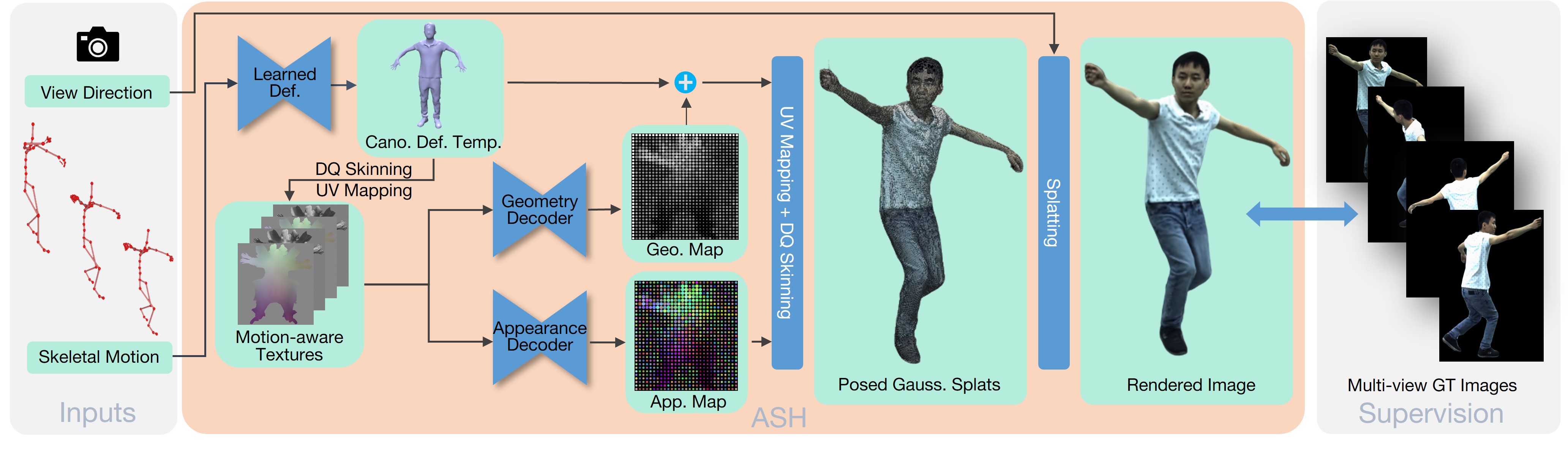}
\vspace{-8pt}
\caption{
    \OURS generates high-fidelity rendering given a skeletal motion and a virtual camera view. 
    A motion-dependent, canonicalized template mesh is generated with a learned deformation network.
    From the canonical template mesh, we can render the motion-aware textures, which are further adopted for predicting the Gaussian splat parameters with two 2D convolutional networks, i.e., the Geometry and Appearance Decoder, as the texels in the 2D texture space.
    Through UV mapping and DQ skinning, we warp the Gaussian splats from the canonical space to the posed space.
    Then, splatting is adopted to render the posed Gaussian splats.
}
\label{fig:pipline}
\end{figure*}

%% file: sec/4_results.tex
%
%
\input{fig/4_gallery}
\section{Results} 
\label{sec:results}

\paragraph{Dataset}
\label{subsec:datset}
We adopted the DynaCap dataset~\cite{habermann2021} to quantitatively and qualitatively assess the effectiveness of our approach.
We selected two representative subjects from the DynaCap dataset wearing loose and tight types of apparel for evaluating the accuracy of novel-view rendering and generalization ability to novel poses.
Following the protocol proposed in DDC~\cite{habermann2021}, we train our model using the training splits from the DynaCap dataset.
Here, we hold out 4 camera views to assess the novel-view rendering accuracy. 
Moreover, we evaluate the model's generalization ability to novel poses with motion sequences from the testing splits.

In addition to the DynaCap dataset, we recorded two novel sequences featuring distinct subjects to showcase the performance of our model qualitatively. 
The recorded subjects perform everyday motions such as dancing, jogging, and jumping.
The sequences are recorded using a calibrated multi-camera system with 120 cameras at a frame rate of 25 fps. 
Separate training and testing sequences are recorded with a duration of 27,000 frames and 7,000 frames, respectively.
All the captured frames are annotated with 3D skeletal poses~\cite{captury} and foreground segmentations~\cite{BMSengupta20,kirillov2023segment}.

%
\subsection{Qualitative Results}
\label{subsec:qualitative}

We evaluate the performance of \OURS on subjects from the DynaCap~\cite{habermann2021} dataset and our newly recorded sequences. 

\noindent\textbf{Novel View Synthesis.} 
Fig.~\ref{fig:gallery} presents the novel view synthesis results rendered from camera views unseen during training. 
\OURS yields photorealistic rendering in real time, capturing sharp wrinkles details and view-dependent appearances. 
Remarkably, it can even generalize to loose types of apparel and faithfully recovers the clothing dynamics, e.g., the swing of the skirts. 

\noindent\textbf{Novel Pose Synthesis.} We further
show the results generated on novel poses extracted from the testing sequences in Fig.~\ref{fig:gallery}. 
Given poses that significantly deviate from the training poses, our method still generates high-quality renderings with motion-aware appearances. 
For the dynamic results, we refer to the supplemental video.
%
%
\subsection{Comparisons}
\label{subsec:comaprison}

\noindent\textbf{Competing Methods.} 
We compare our model with the state of the arts on animatable neural human rendering: 
1) DDC~\cite{habermann2021} features a mesh-based approach where the geometry is represented with a learned embedded graph, and the appearance is encoded using learned dynamic textures. Specifically, DDC is the only real-time approach among competing methods, while other hybrid methods typically take seconds to volume-render an image.
2) TAVA~\cite{li2022tava} is a hybrid approach depicting the shape, appearance, and skinning weights as implicit fields in canonical space. The samples in the posed space are canonicalized w.r.t. the skeleton through iterative root finding.
3) NA~\cite{liu2020neural} conditions the canonical color and density field of dynamic characters on the learned feature texture of the parametric human body models. The canonicalization of spatial samples is achieved by inverse kinematics.
4) HDHumans~\cite{habermann2023hdhumans} models the appearance of dynamic humans as the appearance and density fields conditioned on the feature texture map of the motion-aware deformable template mesh. Notably, the template mesh will deform w.r.t. the implicit density field, improving the alignment between the observation and canonical space.

\noindent\textbf{Metrics.} 
We adopt the Peak Signal-to-Noise Ratio (PSNR) metric to measure the quality of the rendered image.
Moreover, we adopt the learned perceptual image patch similarity (LPIPS) ~\cite{zhang2018perceptual} that better mirrors human perception. 
Note that the metrics are assessed at a 1K resolution, averaged across every 10th frame throughout the sequence. 
Here, we denote the subject with tight outfits as \textit{Tight Outfits}, and the other wearing loose clothing as \textit{Loose Outfits}.

\noindent\textbf{Quantitative Comparison.} 
Tab.~\ref{tab:quantinovelview} and Tab.~\ref{tab:quantinovelpose} illustrate the quantitative comparison against the competing methods on novel view and pose synthesis.
Compared with the real-time capable methods, our method significantly outperforms DDC~\cite{habermann2021} in PSNR and LPIPS regarding novel-view synthesis, highlighting our method's superiority in capturing the motion-aware appearances from the training data.
In novel pose synthesis, compared to DDC~\cite{habermann2021}, our method demonstrates significantly improved performance. 
This underscores our method's generalization ability to novel motions. 
As for the comparison against the non-real-time approaches, our method consistently surpasses previous works regarding PSNR and LPIPS.
Notably, our method is capable of real-time rendering and achieves remarkably better quantitative accuracy than HDHumans in novel-view synthesis, and comparable performances in novel-pose synthesis.

\input{table/comparison_quat_novel_view}
\input{table/comparison_quant_novel_pose}
\input{fig/4_comparison_qual}
\noindent\textbf{Qualitative Comparison.} 
Fig.~\ref{fig:comparison} comprises the qualitative comparison on the novel-view and novel-pose rendering:
TAVA~\cite{li2022tava} struggles to handle 
various 
motions in the DynaCap dataset~\cite{habermann2021}, resulting in blurry renderings.
While NA~\cite{liu2021neural} effectively captures details for subjects wearing tight apparel, it struggles with significant artifacts for subjects in loose outfits. 
This issue arises from the inherent challenge of representing loose clothing as residual displacements on the parametric human body model.
HDHumans~\cite{habermann2023hdhumans} stands out among non-real-time competing methods, producing high-fidelity renderings with sharp details. 
However, due to the extensive sampling needed for volume rendering, it takes seconds for HDHumans to render a single frame. In contrast, \OURS excels by delivering rendering quality that matches or exceeds HDHumans' quality in real-time.

DDC~\cite{habermann2021} is 
the only 
competing method with real-time capability. 
Although it captures coarse motion-aware appearances, 
its output tends to be blurry and lacks detail. 
\OURS matches the real-time capability as DDC, while generating renderings with much finer details.

%
%
\input{table/ablations_quant}
\input{fig/4_ablations_qual}

\subsection{Ablations}
\label{subsec:ablations}

To assess the effectiveness of the major components of our method, we conduct the following ablative experiments on the novel view and motion synthesis tasks.

\noindent\textbf{Motion Conditions.} 
Our method depicts the appearance of the clothed human through motion-aware, deformable Gaussian splats in the canonical space.
To assess the efficacy of the motion conditions, we remove the motion-aware decoder and learn the appearance parameters of Gaussian splats from a truncated training sequence of 1,000 frames, termed as~\textbf{w/o mot.}. 
As seen in Tab.~\ref{tab:supplablation} and Fig.~\ref{fig:ablation}, without motion conditioning, the synthesized results fail to recover the clothing dynamics and suffer from severe artifacts.

\noindent\textbf{Motion-aware Offset.} 
The motion-aware offset is adopted to account for the non-rigid motion-dependent deformation of the Gaussian splats.
We remove the motion-aware offset applied to the canonical Gaussian splats, only allowing the appearance to be motion-dependent, termed as~\textbf{w/o disp.}. 
As shown in Tab.~\ref{tab:supplablation}, excluding the learned motion-aware offset leads to worse quantitative performance and noticeable blurry artifacts on the rendered images.

\noindent\textbf{Texture Resolution.} The animatable Gaussian splats are parameterized as texels in the texture space of the deformable template mesh, where the resolution is set to $256$. 
To study the impact on the resolution of the texture space, we conducted ablative studies with different resolutions, i.e., halved resolution termed as \textbf{w/ 128.res.}, and doubled resolution termed as \textbf{w/ 512.res.}. 
As illustrated in Tab.~\ref{tab:supplablation} and 
Fig.~\ref{fig:ablation}, doubling the resolution results in comparable results, while it significantly increases computational complexity 
in both the U-Net ~\cite{ronneberger2015u} evaluation and tile-based rasterization, preventing the model from being real-time compatible.
On the other hand, reducing the resolution to 128 leads to a significant decline in perceptual metrics and blurry rendering.

As seen in Tab.~\ref{tab:supplablation} and  Fig.~\ref{fig:ablation}, our method outperforms the design alternatives quantitatively and qualitatively. 

%% file: fig/4_gallery.tex
\begin{figure*}[t]
\centering
\includegraphics[width=0.98\textwidth]{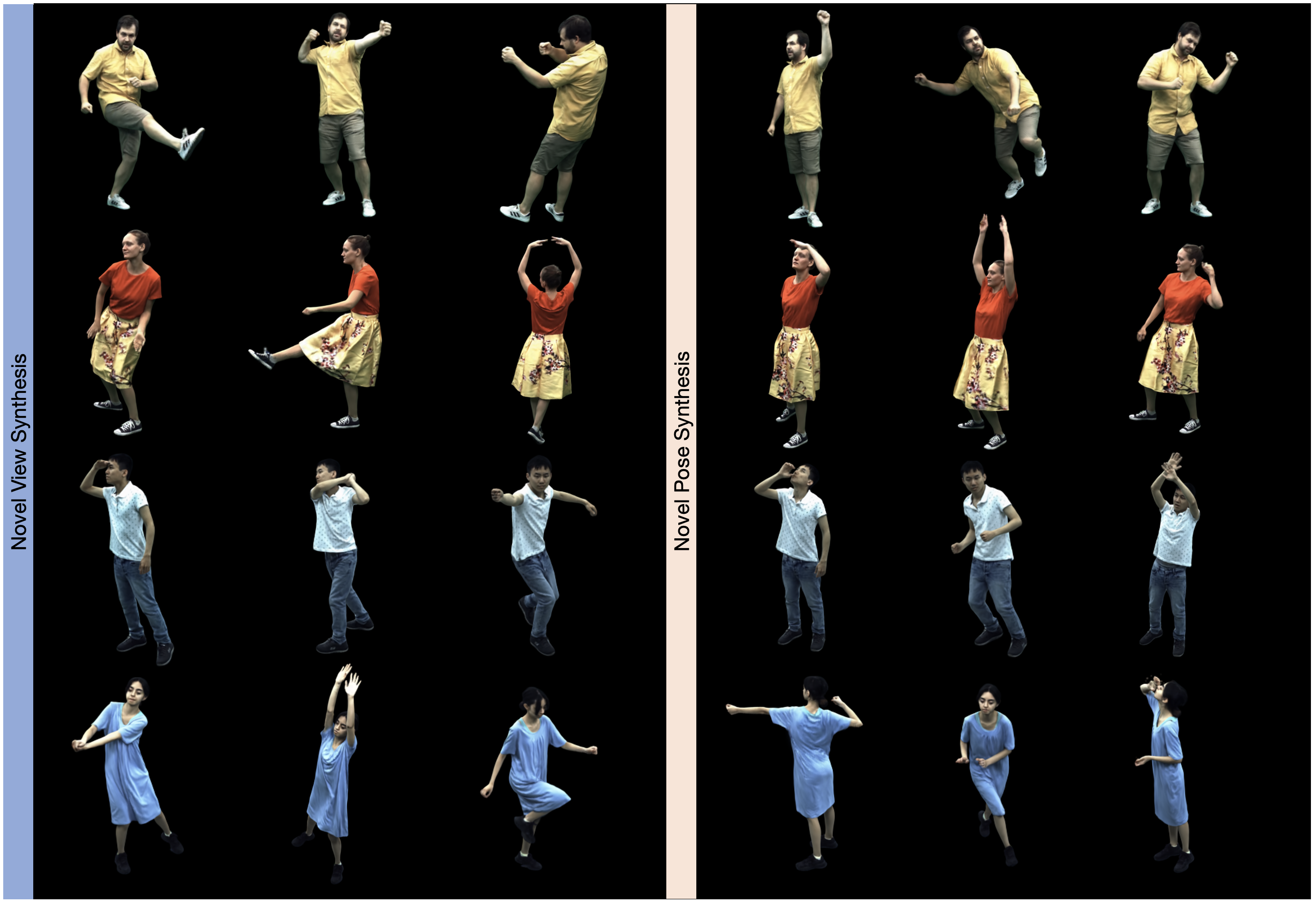}
\vspace{-8pt}
\caption{
    \textbf{Qualitative Results.} 
    We present the results generated with ASH regarding novel view and pose synthesis. 
    Note that our methods can produce high-quality rendering with delicate, motion-aware details for novel views and skeletal motions. 
    }
    \label{fig:gallery}
\end{figure*}

%% file: table/comparison_quat_novel_view.tex
\begin{table}[t]
\small
    \centering
    \begin{tabular}{|l|c|cc|cc|}
    \hline
    \multicolumn{1}{|c|}{}&\multicolumn{1}{c|}{}&\multicolumn{2}{c|}{Tight Outfits}&\multicolumn{2}{c|}{Loose Outfits}\\
    \cline{3-6}
    \multirow{-2}{*}{\textbf{Methods}}  & \multirow{-2}{*}{\textbf{RT}} & \textbf{PSNR}  & \textbf{LPIPS}  & \textbf{PSNR} & \textbf{LPIPS}   \\
    \hline
    TAVA        & \bad      & 24.61          & 62.26          & 27.31         & 37.55 \\
    NA          & \bad      & 30.33          & 23.71          & 25.30         & 50.01 \\
    HDHumans    & \bad      & 30.98          & \cellcolor{antiquebrass} 15.09          & \cellcolor{antiquebrass} 29.24         & \cellcolor{antiquebrass} 15.79 \\
    DDC         & \good     & \cellcolor{antiquebrass}  31.21          & 22.56          & 28.10         & 31.68 \\
    
    \hline
    \textbf{Ours} & \good & \cellcolor{yellow} 35.84  & \cellcolor{yellow} 11.92 & \cellcolor{yellow} 35.47 &  \cellcolor{yellow} 8.30 \\
    \hline
    \end{tabular}
    \caption{
        \textbf{Quantitative Comparison on Novel View Synthesis.} We quantitatively compare \OURS with other methods on seen skeletal motions but unseen views. We highlight the \colorbox{yellow}{best} and  \colorbox{antiquebrass}{second-best} scores.
        We outperform previous real-time and even non-real-time methods in all matrices by a large margin.
    }
    \label{tab:quantinovelview}
\end{table}

%% file: table/comparison_quant_novel_pose.tex
\begin{table}[t]
\small
    \centering
    \begin{tabular}{|l|c|cc|cc|}
    \hline
    \multicolumn{1}{|c|}{}&\multicolumn{1}{c|}{}&\multicolumn{2}{c|}{Tight Outfits}&\multicolumn{2}{c|}{Loose Outfits}\\
    \cline{3-6}
    \multirow{-2}{*}{\textbf{Methods}}     & \multirow{-2}{*}{\textbf{RT}} & \textbf{PSNR}  & \textbf{LPIPS}  & \textbf{PSNR} & \textbf{LPIPS}   \\
    \hline
    TAVA         & \bad & 28.30          & 37.47          & 26.31         & 50.11 \\
    NA           & \bad & \cellcolor{antiquebrass} 28.78          & 25.78          & 25.03         & 44.20 \\
    HDHumans     & \bad & 28.17          & \cellcolor{yellow} 20.69          & \cellcolor{antiquebrass} 26.71         & \cellcolor{antiquebrass} 22.75 \\
    DDC          & \good & 27.77          & 30.16          & 26.43         & 32.22 \\
    
    \hline
    \textbf{Ours} & \good & \cellcolor{yellow} 28.90  &  \cellcolor{antiquebrass} 22.83 & 
    \cellcolor{yellow} 27.12 &  \cellcolor{yellow} 20.22 \\
    \hline
    \end{tabular}
    \caption{\textbf{Quantitative Comparison on Novel Pose Synthesis.} We quantitatively compare \OURS with other methods on unseen skeletal motions and unseen views. \OURS achieves the highest PSNR and the second-best LPIPS on the subject with tight outfits, and outperforms other methods for the subject with loose clothing.
    }
    \label{tab:quantinovelpose}
\end{table}

%% file: fig/4_comparison_qual.tex
\begin{figure*}[t]
\centering
\includegraphics[width=0.99\linewidth]{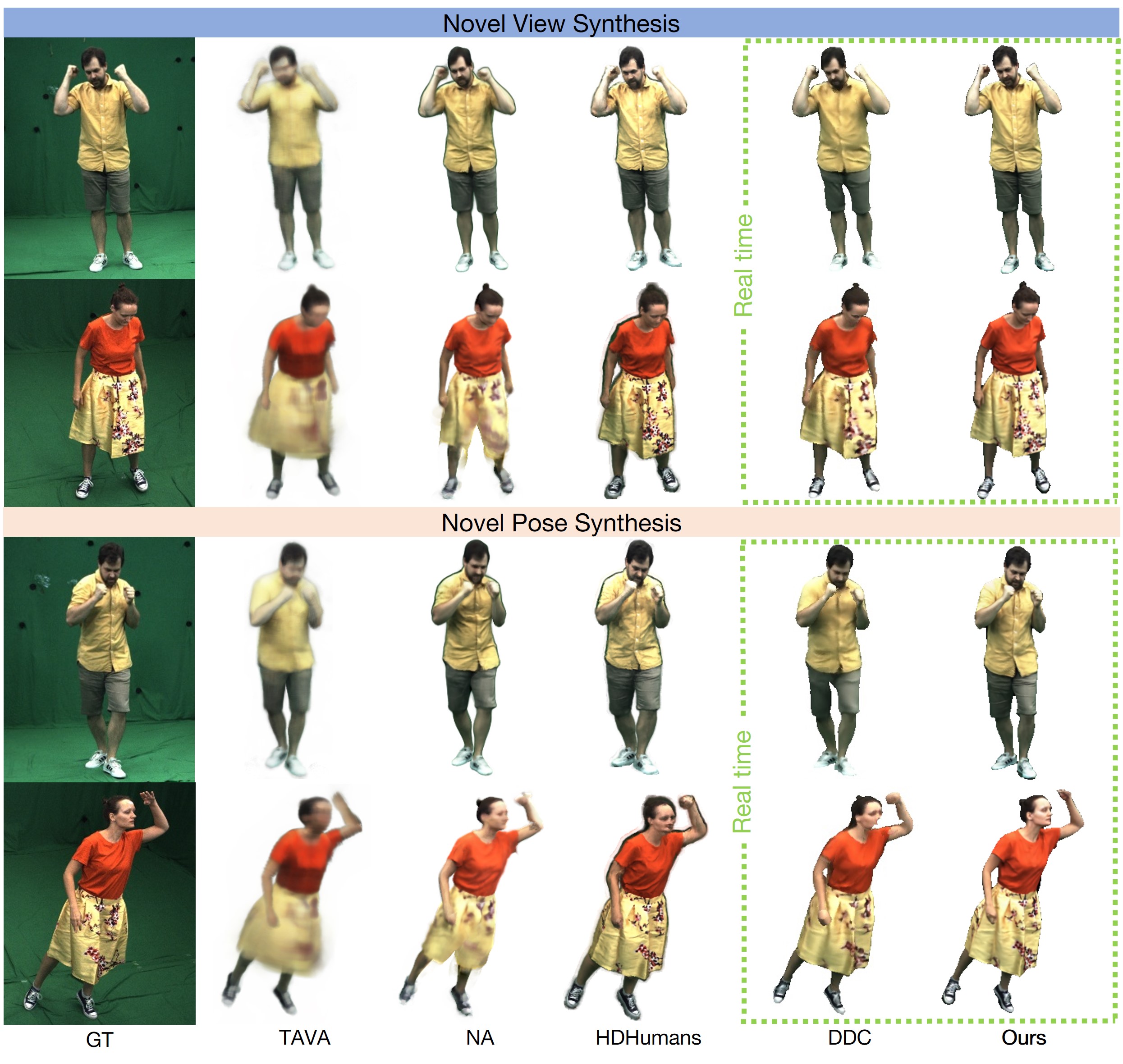}
\vspace{-10pt}
\caption{
    \textbf{Qualitative Comparison}. We compared our methods with the state of the arts, i.e., TAVA~\cite{li2022tava}, NA~\cite{liu2021neural}, HDHumans~\cite{habermann2023hdhumans}, DDC~\cite{habermann2021}, in novel view and novel motion synthesis. Note that our results significantly 
    outperform the real-time methods in quality while showing comparable or even better results than the offline methods.
}
\label{fig:comparison}
\end{figure*}

%% file: table/ablations_quant.tex
\begin{table}[t]
\small
    \centering
    \begin{tabular}{|l|c|c|c|c|}
    \hline
    &\multicolumn{2}{c|}{Training Pose}&\multicolumn{2}{c|}{Testing Pose} \\
    \cline{2-5}
     \multirow{-2}{*}{\textbf{Methods}}& \textbf{PSNR}  & \textbf{LPIPS} & \textbf{PSNR}  & \textbf{LPIPS} \\
    \hline
     w/o mot.       & 27.19             & 33.16          & 26.86         & 32.35  \\
     w/o disp.       & 33.21            & 17.34           & \cellcolor{yellow} 27.33         & 25.32  \\
     w/ 128.res.     & 35.15            & 11.26           & 27.13         & 22.00  \\
     w/ 512.res.     & \cellcolor{antiquebrass} 35.28            & \cellcolor{antiquebrass} 8.60            & 27.00         & \cellcolor{antiquebrass} 21.13  \\
    
    \hline
    \textbf{Ours}   & \cellcolor{yellow} 35.47            & \cellcolor{yellow} 8.30            & \cellcolor{antiquebrass}27.13     & \cellcolor{yellow} 20.22 \\
    \hline
    \end{tabular}
    \caption{\textbf{Ablation Study}. 
        We assess our design choices on the image synthesis tasks on the subject with loose outfits. 
        Our method achieves better performance against the design alternatives.
    }
    \label{tab:supplablation}
\end{table}

%% file: fig/4_ablations_qual.tex
\begin{figure}[t]
\centering
\includegraphics[width=0.98\linewidth]{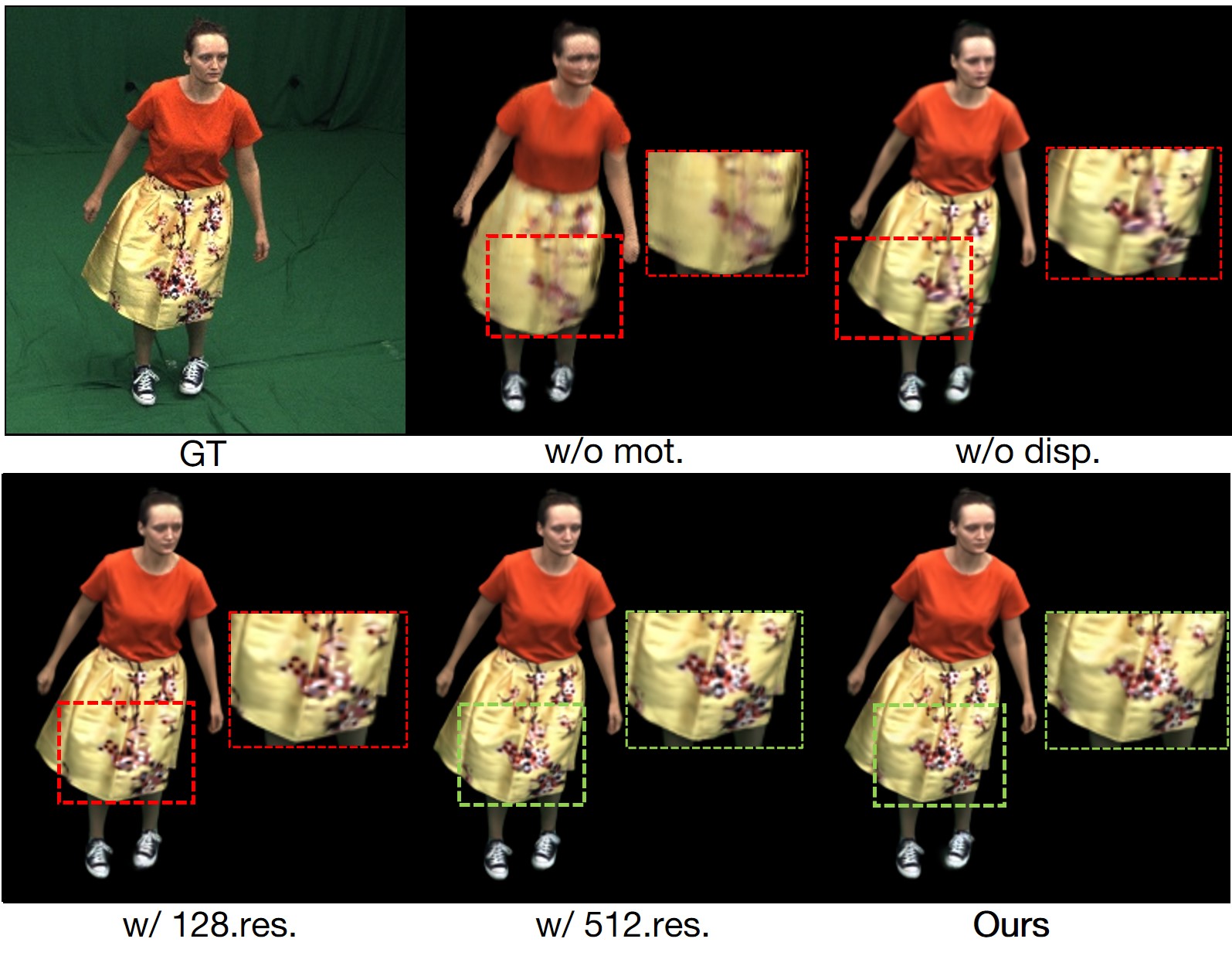}
\caption{
    \textbf{Qualitative Ablation.}         
    We compare our design choices on the image synthesis task. 
    Our method excels in rendering quality and detail recovery.
    Our method shows compatible rendering quality as \textbf{w/ 512.res.} with doubled texel resolution and much sharper rendering than \textbf{w/ 128.res.} with halved texel resolution.
}
\label{fig:ablation}
\end{figure}

%% file: sec/6_conclusion.tex
%
%
\section{Conclusion} 
\label{sec:conclusion}
In this paper, we introduce \OURS, a real-time method for high-quality rendering of animated humans, learned solely from multi-view videos.
\OURS attaches the 3D Gaussians splats, initially designed for static scenes, onto a deformable mesh template. 
Bridged by the mesh's UV parameterization, we can efficiently learn the 3D Gaussians in 2D texture space as an image-2-image translation task.
\OURS demonstrates significantly better performances quantitatively and qualitatively than state-of-the-art, real-time capable methods on animatable human rendering, and even better performance than the state-of-the-art offline methods.
Currently, \OURS does not update the underlying deformable template mesh. In the future, we will explore whether the Gaussian splatting can directly improve the 3D mesh geometry.

%% file: sec/7_ack.tex
\section{Acknowledgement} 
\label{sec:ack}
Christian Theobalt was supported by ERC Consolidator Grant 4DReply (No.770784). 
Adam Kortylewski was supported by the German Science Foundation (No.468670075). 
This project was also supported by the Saarbrucken Research Center for Visual Computing, Interaction, and AI.

%% file: suppl_sec/1_ddc.tex
\section{Implementation Details}
\label{suppl:ddc}

In the main paper, we mentioned that \OURS learns the Gaussian splat parameters in the 2D texture space of an animatable human template $M(\boldsymbol{\theta}_{f})=\mathbf{V}_{f}$.
Here, we provide more details regarding the deformable template mesh and the motion-aware decoders.

\noindent \textbf{Deformable Template Mesh.} We adopt the formulation introduced in Habermann et al.~\cite{habermann2021} for modeling the deformable template mesh, which deforms the template mesh vertices $\mathbf{V}_{\mathrm{m}}$ in the canonical space with a learned embedded deformation~\cite{embedded,sorkine2007rigid}:
\begin{equation} 
\label{eq:ddc_deform}
\bar{\mathbf{V}}_{f,i}=\mathbf{D}_{i}+\sum_{j \in \mathcal{N}_{\mathrm{nv},i}}w_{i,j}(R(\mathbf{A}_j)(\mathbf{V}_{\mathrm{m}, i}-\mathbf{V}_{\mathrm{G},j}) + \mathbf{V}_{\mathrm{G},j}+\mathbf{T}_{j})
\end{equation}
where $\bar{\mathbf{V}}_{f,i} \in \mathbb{R}^{3}$ denotes the deformed template vertices in the rest pose. 
$\mathcal{N}_{\mathrm{nv},i} \in \mathbb{N}$ indicates the indices for the embedded graph node~\cite{embedded} that are connected to the $i$ th vertex on the template mesh.
$\mathbf{V}_{\mathrm{G},j} \in \mathbb{R}^{3}$, $\mathbf{A}_j \in \mathbb{R}^{3}$, and $\mathbf{T}_j\in \mathbb{R}^{3}$ denotes the rest positions, Euler angles, and translations of the embedded graph nodes. 
Notably, the connectivity of the embedded graph $\mathbf{V}_{\mathrm{G},j}$ can be adopted by simplifying the template mesh $M$ using quadric edge collapse decimation~\cite{cignoni2011meshlab,garland1997surface}.
Moreover, the connection, as well as the connection weights $w_{i,j}$, between the template mesh $\mathbf{V}_{\mathrm{m}}$ and the embedded graph are generated Sumner et al.~\cite{embedded}.
$R(\cdot) \in \mathbb{R}^{3\times3}$ denotes the function that converts the Euler angle to a rotation matrix.
$\mathbf{D}_i \in \mathbb{R}^{3}$ indicates the per-vertex displacement to model an even finer level of geometry details. 
Specifically, embedded graph parameters $\mathbf{V}_{\mathrm{G},j}$, $\mathbf{A}_j$, and  per-vertex displacements $\mathbf{D}_i$ are derived from skeletal motion $\boldsymbol{\bar{\theta}}_{f}$  with structure-aware graph convolution neural networks. We refer to Harbermann et al.~\cite{habermann2021} for more details.

\noindent\textbf{Motion-aware Decoders.} \OURS adopts motion-aware 2D convolutional neural networks, i.e., the geometry network $\mathcal{E}_{\mathrm{geo}}$, and the appearance network $\mathcal{E}_{\mathrm{app}}$, predicting the geometry and appearance parameters of the Gaussian splats from the motion-aware textures $(\mathbf{T}_{\mathrm{n},f}, \mathbf{T}_{\mathrm{p},f})$. 
Both the geometry network $\mathcal{E}_{\mathrm{geo}}$ and the appearance network $\mathcal{E}_{\mathrm{app}}$ are U-Nets implemented following the configuration mentioned in Olaf et al.~\cite{ronneberger2015u}. 
Specifically, we channel-wise concatenate the global appearance features $\Phi_f \in \mathbb{R}^{16}$ to the bottleneck features of the appearance network $\mathcal{E}_{\mathrm{app}}$ to account for the lighting variations in the studio. 
The global appearance feature $\Phi_f$ is derived from positional encoded skeleton root translation with a 3-layer shallow MLP, of which the width is set to $32$.

%% file: suppl_sec/2_ablations.tex
\section{Ablations}
\label{suppl:ablations}
In this section, we provide more ablative studies to demonstrate the effectiveness of \OURS.

\noindent\textbf{Number of Camera Views.} To assess the robustness of \OURS against sparser camera view supervision, we conducted ablative experiments that take multi-view videos from $12$, $30$, and $60$ cameras as supervision, termed as \textbf{w/ 12.cam}, \textbf{w/ 30.cam}, and  \textbf{w/ 60.cam}. 
Note that the selected camera views are evenly distributed in the studio.
As illustrated in Fig.~\ref{fig:supplablation} and Tab.~\ref{tab:ablation2}, \OURS can still accurately synthesize the animatable characters when training with sparser input views.

\noindent\textbf{The Impact of 2D Learning.} 
To validate the efficacy of the 2D texel paradigm for 3D Gaussian splats, we conducted an ablative experiment that predicts the 3D Gaussian parameters directly from 3D, termed as \textbf{w/ MLP}.
We adopted an 8-layer MLP that consumes the skeletal motion $\boldsymbol{\bar{\theta}}_{f}$ and the positional-encoded canonical Gaussian position $\boldsymbol{\bar{\mu}}_i$, predicting the Gaussian splat parameters $\{\mathcal{G}_{i}\}$ in the canonical space.
Specifically, the width for the hidden layers of the MLP is set to $256$. 
Similar to \OURS, the canonical Gaussian splats are transformed to observation space through Dual Quaternion skinning~\cite{kavan2007skinning}.
As illustrated in Fig.~\ref{fig:supplablation} and Tab.~\ref{tab:ablation2}, directly learning the Gaussian parameters in 3D will lead to blurry rendering and cannot preserve the motion-dependent wrinkle details. 
In contrast, \OURS, which formulates the learning of 3D Gaussian splats as image translation in 2D texel space, delivers high-quality rendering with delicate details.

\input{table/ablations_supp}
\input{fig/suppl_ablation}

\noindent \textbf{The Reliance on the Accuracy of the Template. } 
Although our method is conditioned on a template mesh, it can compensate for tracking errors with learnable motion-aware residual deformations for the Gaussian splats.
To validate our method's robustness against errors in mesh tracking, we replaced the original template mesh with \textit{SMPL body meshes}~\cite{loper2015smpl}. 
Despite large deviations between the template and "the real surface," our method generates visually plausible results (Fig.~\ref{fig:smpl}) and achieves significantly better quantitative performance than the SOTA real-time methods (Tab.~\ref{tab:smpl}), which heavily relies on an accurate deformable template~\cite{habermann2021}.

\input{table_rebuttal/rebuttal_ablation_quant}
\input{figure_rebuttal/fig_smpl}

%% file: table/ablations_supp.tex
\begin{table}[t]
\small
    \centering
    \begin{tabular}{|l|c|c|c|c|}
    \hline
    &\multicolumn{2}{c|}{Training Pose}&\multicolumn{2}{c|}{Testing Pose} \\
    \cline{2-5}
     \multirow{-2}{*}{\textbf{Methods}}& \textbf{PSNR}  & \textbf{LPIPS} & \textbf{PSNR}  & \textbf{LPIPS} \\
    \hline
     w/ MLP         & 28.05         & 34.93         & 26.79         & 35.00 \\
     w/ 12.cam      & 33.50         & 10.76         & 26.96         & 21.66 \\
     w/ 30.cam      & 35.37         & 8.52          & 27.08         & 20.76 \\
     w/ 60.cam      & \cellcolor{yellow} 35.49         & \cellcolor{antiquebrass} 8.32          & \cellcolor{yellow} 27.14         & \cellcolor{antiquebrass} 20.42 \\
    \hline
    \textbf{Ours}   & \cellcolor{antiquebrass} 35.47         & \cellcolor{yellow} 8.30          & \cellcolor{antiquebrass} 27.13         & \cellcolor{yellow} 20.22 \\
    \hline
    \end{tabular}
    \caption{\textbf{Ablation Study}. 
        We further assess our design choices on the image synthesis tasks with the subject wearing loose outfits in the DynaCap~\cite{habermann2021} dataset. 
        We highlight the \colorbox{yellow}{best} and the \colorbox{antiquebrass}{second-best} scores.
    }
    \label{tab:ablation2}
    \vspace{-1.5em}
\end{table}

%% file: fig/suppl_ablation.tex
\begin{figure}[t]
\centering
\includegraphics[width=0.98\linewidth]{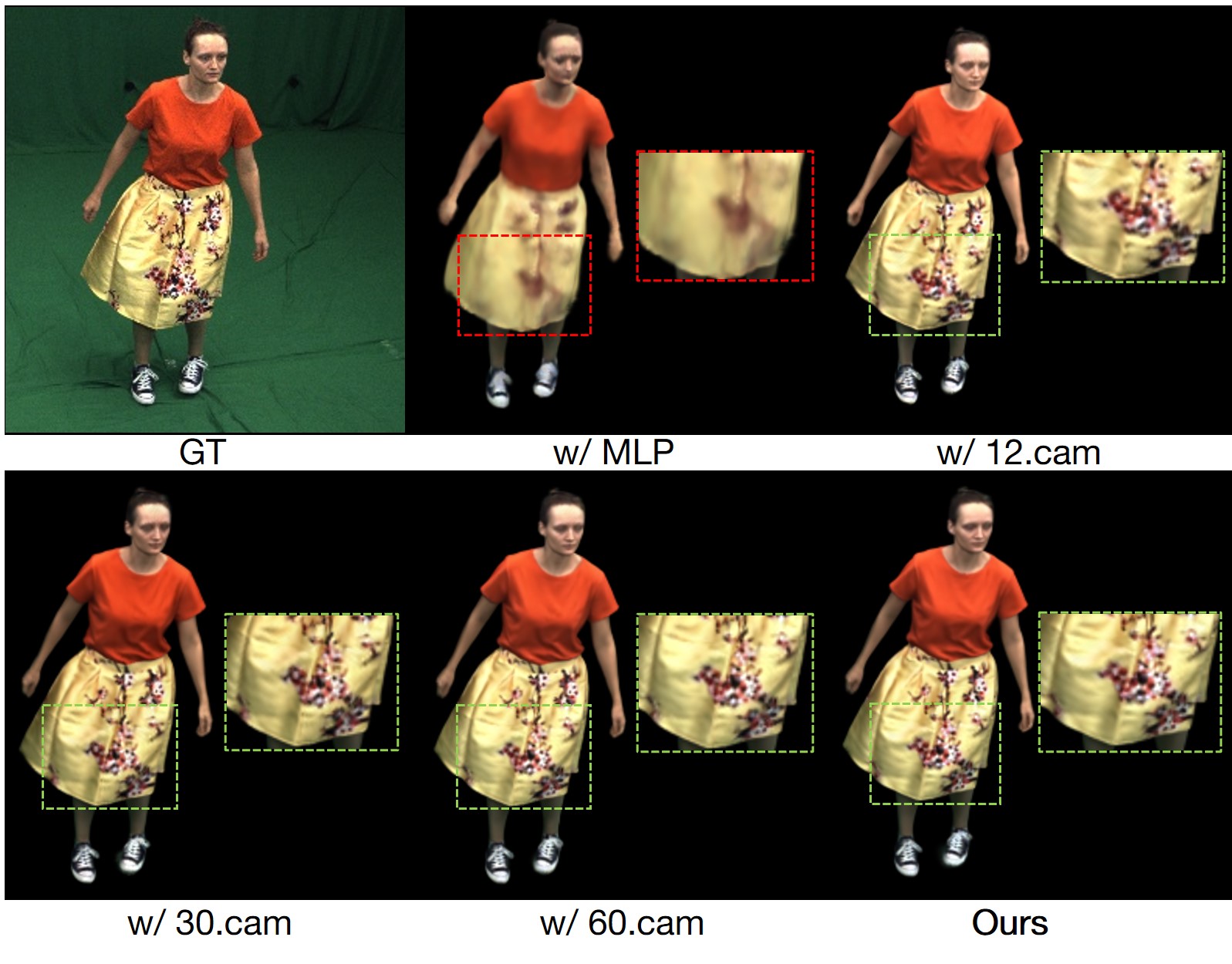}
\caption{
    \textbf{Qualitative Ablation.} 
    We compare \OURS with the models that take alternative design choices. 
    \OURS excels in rendering quality than the model directly learns the Gaussian parameters from 3D canonical space (\textbf{w/ MLP}). 
    Moreover, \OURS exhibits robustness against less training views (\textbf{w/ 12.cam}, \textbf{w/ 30.cam}, \textbf{w/ 60.cam}).
}
\label{fig:supplablation}
\end{figure}

%% file: table_rebuttal/rebuttal_ablation_quant.tex
\begin{table}[t]
\small
    \centering
    \resizebox{!}{25pt}{
    \begin{tabular}{|l|c|c|c|c|}
    \hline
    &\multicolumn{2}{c|}{Training Pose}&\multicolumn{2}{c|}{Testing Pose} \\
    \cline{2-5}
     \multirow{-2}{*}{\textbf{Methods}}& \textbf{PSNR}  & \textbf{LPIPS} & \textbf{PSNR}  & \textbf{LPIPS} \\
    \hline
     SMPL       & 34.93             & 14.90         & 28.75         & 24.87  \\
    \hline
     Ours       & 35.47            & 8.30           & 27.13         & 20.22  \\
    \hline
    \end{tabular}
    }
    \caption{\textbf{ASH conditioned on SMPL.} ASH achieves significantly better quantitative performance than the SOTA real-time methods.
    }
    \label{tab:smpl}
\end{table}

%% file: figure_rebuttal/fig_smpl.tex
\begin{figure}[t]
\centering
\includegraphics[width=1.0\linewidth]{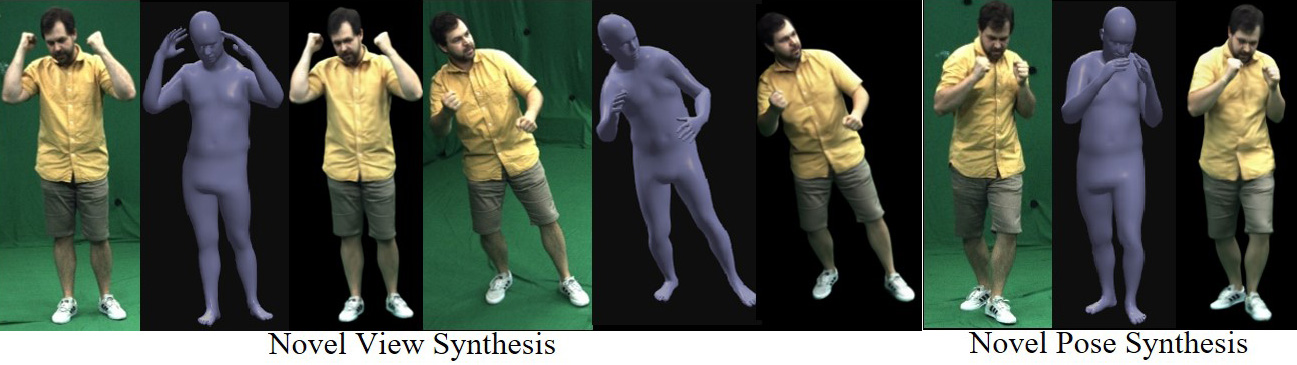}
\caption{
    \textbf{ASH conditioned on SMPL.} Despite large deviations between the underlying template and the real surface, ASH generates visually plausible results.
}
\label{fig:smpl}
\end{figure}

%% file: suppl_sec/3_results.tex
\section{More Results}
\label{suppl:results}
In Tab.~2 in the main paper, we report the quantitative and qualitative performance on the testing set of the DynaCap dataset, which is an established and challenging benchmark, with the testing set containing more than $7000$ frames showing strongly varying poses.

To further highlight the pose generalization ability of ASH, we retarget our skeleton to SMPL motions from the AMASS dataset (DanceDB)~\cite{AMASS:ICCV:2019} to drive our character. 
Fig.~\ref{fig:amass} illustrates that even for motions from an entirely different dataset, ASH could generate photoreal rendering with delicated wrinkle details.

\input{figure_rebuttal/fig_amass}

%% file: figure_rebuttal/fig_amass.tex
\begin{figure}[t]
\centering
\includegraphics[width=1.0\linewidth]{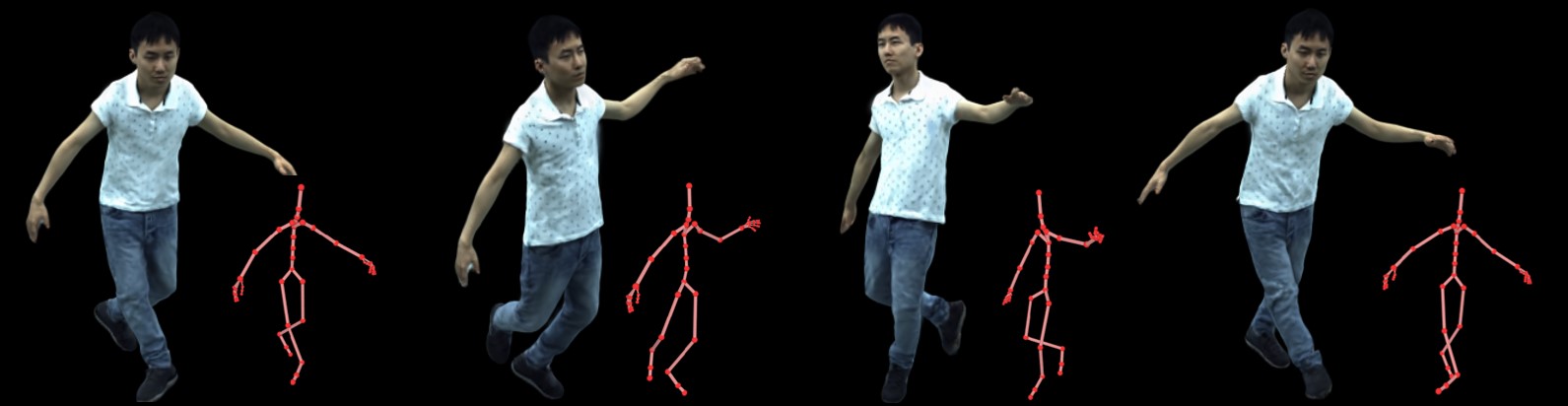}
\caption{
    \textbf{Results with AMASS DanceDB motion.} 
    ASH produces photorealistic rendering given the motion from an entirely different dataset.
}
\label{fig:amass}
\end{figure}

%% file: suppl_sec/4_runtime.tex
\section{Runtime Analysis}
\label{suppl:runtime}

\input{table/runtime}

In this section, we conduct a detailed runtime analysis for each major component in \OURS. 
Specifically, we record the runtime for each major component when rendering a 1K ($1285 \times 940$) image on a single Nvidia Tesla A100 graphics device.
Additionally, the runtime analysis is benchmarked on models with different texture space resolution, specifically at $128$, $512$, and $256$, referred to as \textbf{w/ 128.res}, \textbf{w/ 512.res}, and \textbf{Ours}, respectively.
Here, we divide the rendering pipeline of \OURS into four steps:
\begin{itemize}
    \item Creating the deformable template meshes $M(\boldsymbol{\theta}_{f})$ from skeletal motions $\boldsymbol{\theta}_{f}$ with structure-aware graph convolution networks, termed as \textbf{Stg.1}.
    \item Computing motion-aware texture maps $(\mathbf{T}_{\mathrm{n},f}, \mathbf{T}_{\mathrm{p},f})$ from deformable template meshes $M(\boldsymbol{\theta}_{f})$, termed as \textbf{Stg.2}.
    \item Predicting the canonical Gaussian splats $\{\mathcal{G}_i\}$ with motion-ware geometry decoder $\mathcal{E}_{\mathrm{geo}}$ and appearance decoder $\mathcal{E}_{\mathrm{app}}$, termed as \textbf{Stg.3}.
    \item Performing tile-based rasterization with the predicted Gaussian splats   $\{\mathcal{G}_i\}$, termed as \textbf{Stg.4}. 
\end{itemize}

Tab.~\ref{tab:runtime} illustrates the runtime for each component in \OURS for models with different 2D texel resolutions. 
While halving the texel resolution (\textbf{w/ 128.res.}) speeds up the image synthesis of the animatable humans, it may produce blurry details in the rendered images.
Doubling the texel resolution (\textbf{w/ 512.res.}) results in comparable rendering quality. Nevertheless, it significantly increases computational complexity, preventing the model from being real-time compatible.
In contrast, \OURS can generate high-fidelity renderings of animatable characters in a real-time frame rate.

%% file: table/runtime.tex
\begin{table}[t]
\small
    \centering
    \begin{tabular}{|l|c|c|c|c|c|c|}
    \hline
     {\textbf{Method}}& \textbf{Stg.1}  & \textbf{Stg.2}  & \textbf{Stg.3} & \textbf{Stg.4} & \textbf{Time}  & \textbf{FPS}  \\
    \hline
     w/ 128.res.     & 20.28            & 1.93           & 5.57       & 1.91    & 29.69  & 33.68  \\
     w/ 512.res.     & 24.25            & 16.34          & 18.45       & 4.31    & 63.35   & 15.79 \\
    \hline
    \textbf{Ours}   & \textbf{21.03}    & \textbf{3.60}           & \textbf{7.00}     & \textbf{2.11}     & \textbf{33.74}  & \textbf{29.64} \\
    \hline
    \end{tabular}
    \caption{\textbf{Runtime Analysis}. 
        We present detailed runtime for each major component in \OURS measured in milliseconds. 
        We also report the runtime of the models that take halved and doubled texel resolution, termed as \textbf{w/ 128.res.} and \textbf{w/ 512.res.}, respectively.
        Note that \OURS can render high-quality animatable humans in a real-time frame rate.
    }
    \label{tab:runtime}
    \vspace{-0.5em}
\end{table}

%% file: suppl_sec/6_application.tex
\section{Application}
\label{suppl:application}
In this section, we introduce ASH Player, a real-time application built upon \OURS.

Fig.~\ref{fig:system} presents a screenshot of ASH Player, which runs in the web browser on a personal computer. 
The backend model of ASH Player, i.e., \OURS, is deployed on the GPU cluster server. 
Once users specify the skeletal poses and virtual camera views, ASH Player will present the photoreal rendering of animatable characters, which is real-time computed and streamed from the GPU cluster server. 
Moreover, ASH Player allows users to inspect the animatable characters with spiral camera views.
Please refer to the supplementary video for a more comprehensive visualization.

\input{fig/suppl_system}

%% file: fig/suppl_system.tex
\begin{figure}[t]
\centering
\includegraphics[width=0.98\linewidth]{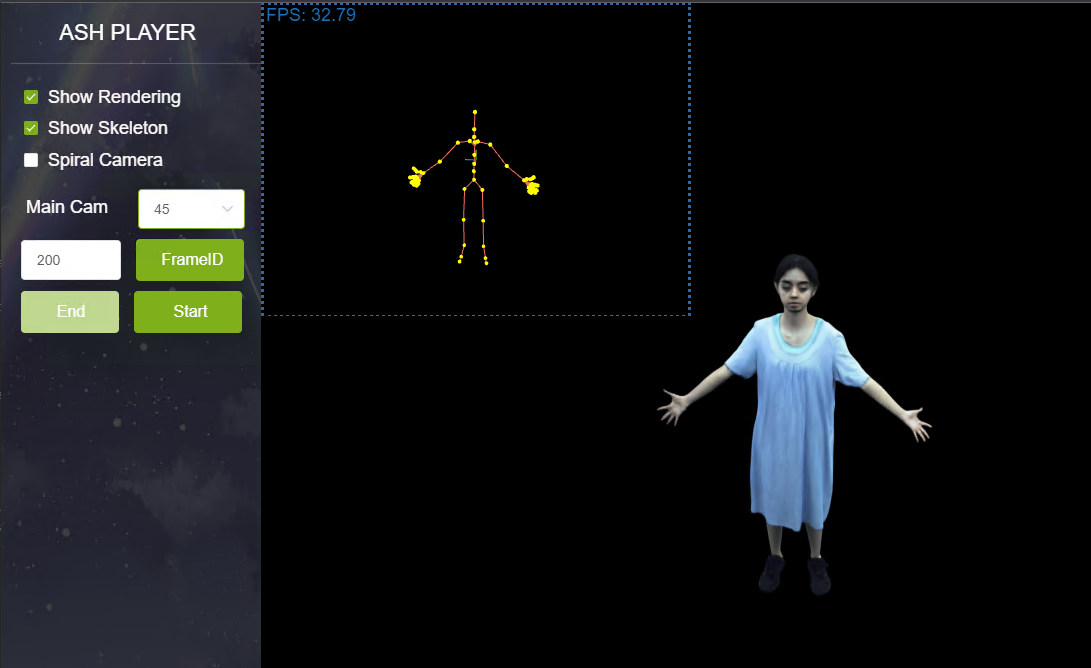}
\caption{
    \textbf{System Overview.} 
    ASH Player is an interface runs in the browser, visualizing the imagery and skeletal poses of animatable characters. 
    The renderings of the animatable humans are computed in real time from the GPU cluster server and streamed to the ASH Player front-end interface on a personal computer.
}
\label{fig:system}
\vspace{-0.5em}
\end{figure}

%% file: suppl_sec/5_limitations.tex
\section{Limitations}
\label{suppl:limitations}

Although \OURS enables high-fidelity, real-time rendering of animatable human characters, it has certain limitations that we hope to address in the future. Firstly, \OURS does not extract detailed explicit geometry from the Gaussian splats. 
We will explore refining the explicit template meshes by backpropagating the gradient from image space into the template meshes using splatting.
Additionally, \OURS does not model topological changes like opening a jacket. 
Future research might focus on modeling the topological changes with the adaptive adding and removal of Gaussian splats introduced in the original 3D Gaussian splatting paper~\cite{kerbl20233d}.
Lastly, as various factors could affect the appearance of dynamic clothed humans, it is unfeasible to establish a one-to-one correspondence between the skeletal motions and the dynamic clothed human appearance. 
Future research will explore different types of fine-grained control to define human rendering, e.g., the external physical forces.